\documentclass{article}

\usepackage{arxiv}
\usepackage{color}
\usepackage[utf8]{inputenc} % allow utf-8 input
\usepackage[T1]{fontenc}    % use 8-bit T1 fonts
\usepackage{hyperref}       % hyperlinks
\usepackage{url}            % simple URL typesetting
\usepackage{booktabs}       % professional-quality tables
\usepackage{amsfonts}       % blackboard math symbols
\usepackage{nicefrac}       % compact symbols for 1/2, etc.
\usepackage{microtype}      % microtypography
\usepackage{cleveref}       % smart cross-referencing
\usepackage{lipsum}         % Can be removed after putting your text content
\usepackage{graphicx}
\usepackage{natbib}
\usepackage{doi}

\title{In-Context LoRA for Diffusion Transformers}

\date{}

\author{
Lianghua Huang
\And
Wei Wang
\And
Zhi-Fan Wu
\AND
Yupeng Shi
\And
Huanzhang Dou
\And
Chen Liang
\And
Yutong Feng
\And
Yu Liu
\And
Jingren Zhou
\AND
\\
{\large Tongyi Lab}\thanks{Emails: Lianghua Huang, Wei Wang, Zhi-Fan Wu, Yutong Feng, Yu Liu, Jingren Zhou \{xuangen.hlh, ww413411, wuzhifan.wzf, fengyutong.fyt, ly103369, jingren.zhou\}@alibaba-inc.com, and Yupeng Shi (shiyupeng.syp@taobao.com). Huanzhang Dou (hzdou@zju.edu.cn, Zhejiang University) and Chen Liang (liangchen2022@ia.ac.cn, Institute of Automation, Chinese Academy of Sciences) contributed to this work during internships at Tongyi Lab.}
}

\renewcommand{\undertitle}{Technical Report}
\hypersetup{
pdftitle={A template for the arxiv style},
pdfsubject={q-bio.NC, q-bio.QM},
pdfauthor={David S.~Hippocampus, Elias D.~Striatum},
pdfkeywords={First keyword, Second keyword, More},
}

\begin{document}
\maketitle

\begin{abstract}
Recent research \citep{huang2024group} has explored the use of diffusion transformers (DiTs) for task-agnostic image generation by simply concatenating attention tokens across images. However, despite substantial computational resources, the fidelity of the generated images remains suboptimal. In this study, we reevaluate and streamline this framework by hypothesizing that \textbf{text-to-image DiTs inherently possess in-context generation capabilities}, requiring only minimal tuning to activate them. Through diverse task experiments, we qualitatively demonstrate that existing text-to-image DiTs can effectively perform in-context generation without any tuning. Building on this insight, we propose a remarkably simple pipeline to leverage the in-context abilities of DiTs: (1) concatenate images instead of tokens, (2) perform joint captioning of multiple images, and (3) apply task-specific LoRA tuning using small datasets (\textit{e.g.,} $20\sim 100$ samples) instead of full-parameter tuning with large datasets. We name our models In-Context LoRA (IC-LoRA). This approach requires no modifications to the original DiT models, only changes to the training data. Remarkably, our pipeline generates high-fidelity image sets that better adhere to prompts. While task-specific in terms of tuning data, our framework remains task-agnostic in architecture and pipeline, offering a powerful tool for the community and providing valuable insights for further research on product-level task-agnostic generation systems. We release our code, data, and models at \url{https://github.com/ali-vilab/In-Context-LoRA}.
\end{abstract}

% keywords can be removed

\begin{figure}
    \centering
    \vspace{-12pt}

    \makebox[\textwidth]{%
        \hspace{-50pt}
        % \fbox{\rule{0pt}{0.95\textheight}\rule{1.1\textwidth}{0pt}} % 用于占位的矩形框
        \includegraphics[width=1.1\textwidth]{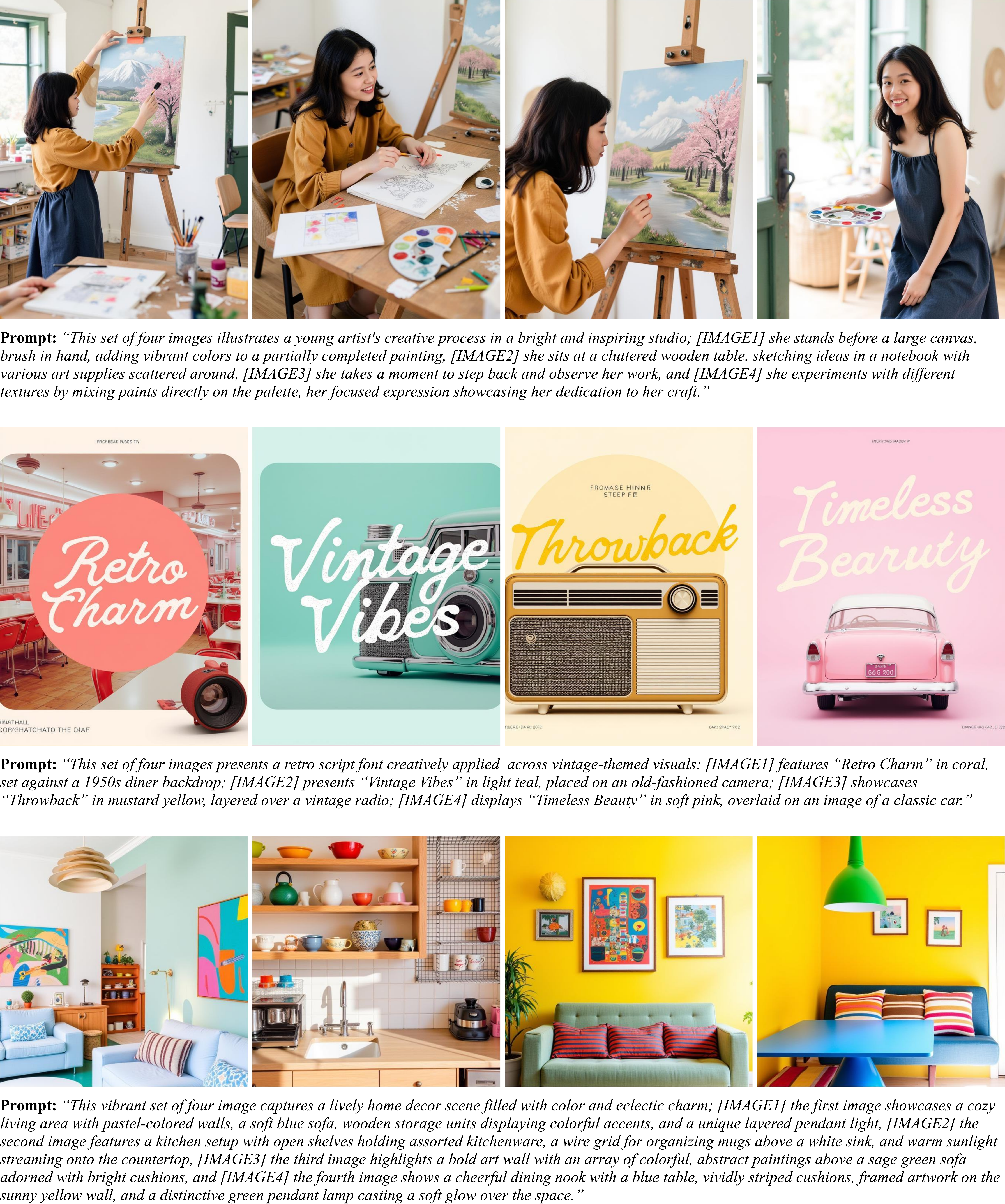}
        \hspace{-50pt}
    }

    \caption{\textbf{In-Context LoRA Generation Examples.} Three tasks from top to bottom: \textit{portrait photography, font design}, and \textit{home decoration}. For each task, four images are generated simultaneously within a single diffusion process using In-Context LoRA models that are tuned specifically for each task.}
    \label{fig:fig1}
\end{figure}

\begin{figure}
    \centering
    \vspace{-12pt}

    \makebox[\textwidth]{%
        \hspace{-50pt}
        % \fbox{\rule{0pt}{0.95\textheight}\rule{1.1\textwidth}{0pt}} % 用于占位的矩形框
        \includegraphics[width=1.1\textwidth]{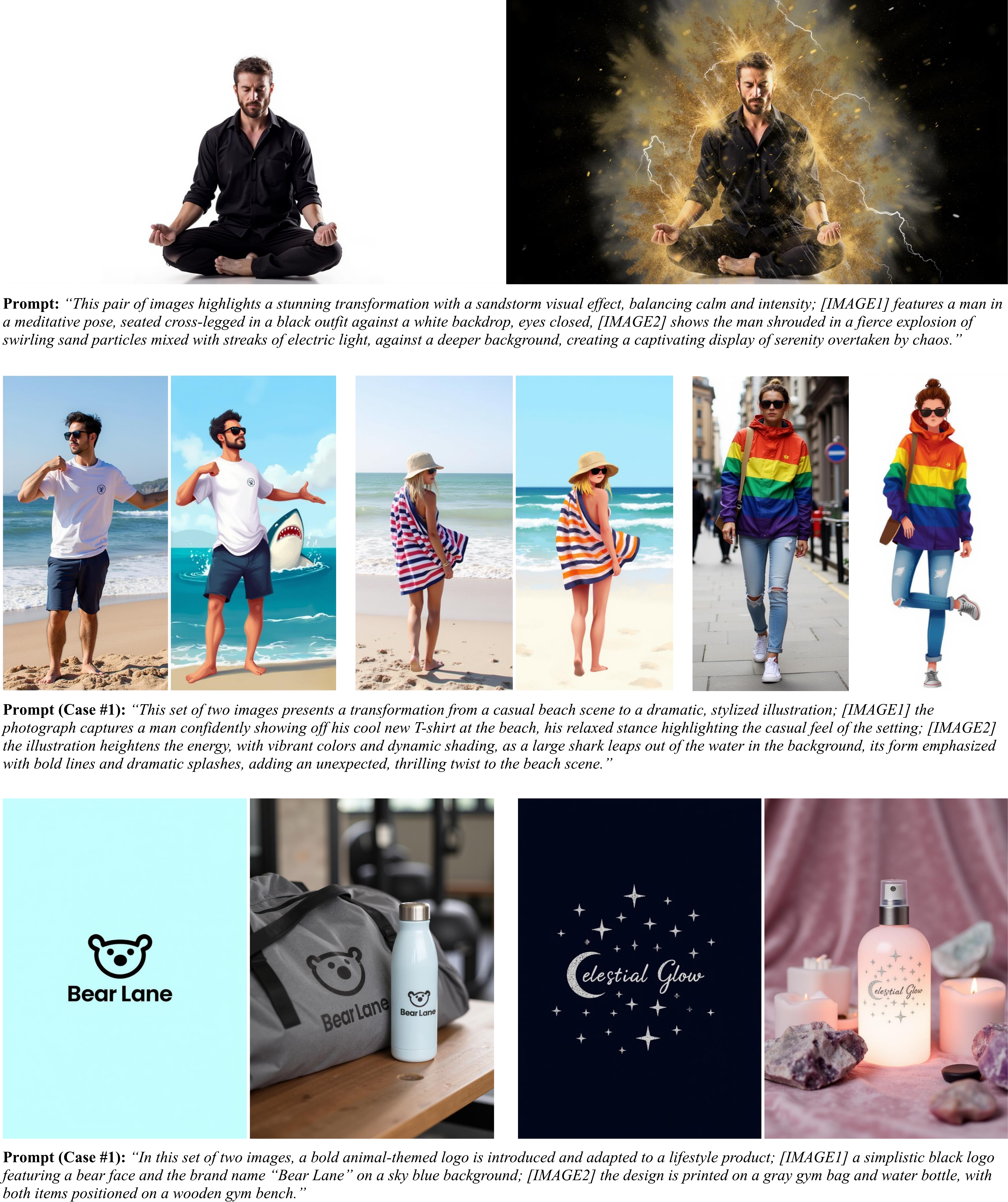}
        \hspace{-50pt}
    }

    \caption{\textbf{In-Context LoRA Generation Examples.} Three tasks from top to bottom: \textit{sandstorm visual effect, portrait illustration}, and \textit{visual identity design}. For each task, an image pair is generated simultaneously within a single diffusion process. The further application of SDEdit for image-conditional generation will be discussed later in the paper.}
    \label{fig:fig2}
\end{figure}

\begin{figure}
    \centering
    \vspace{-12pt}
    
    \includegraphics[width=\textwidth]{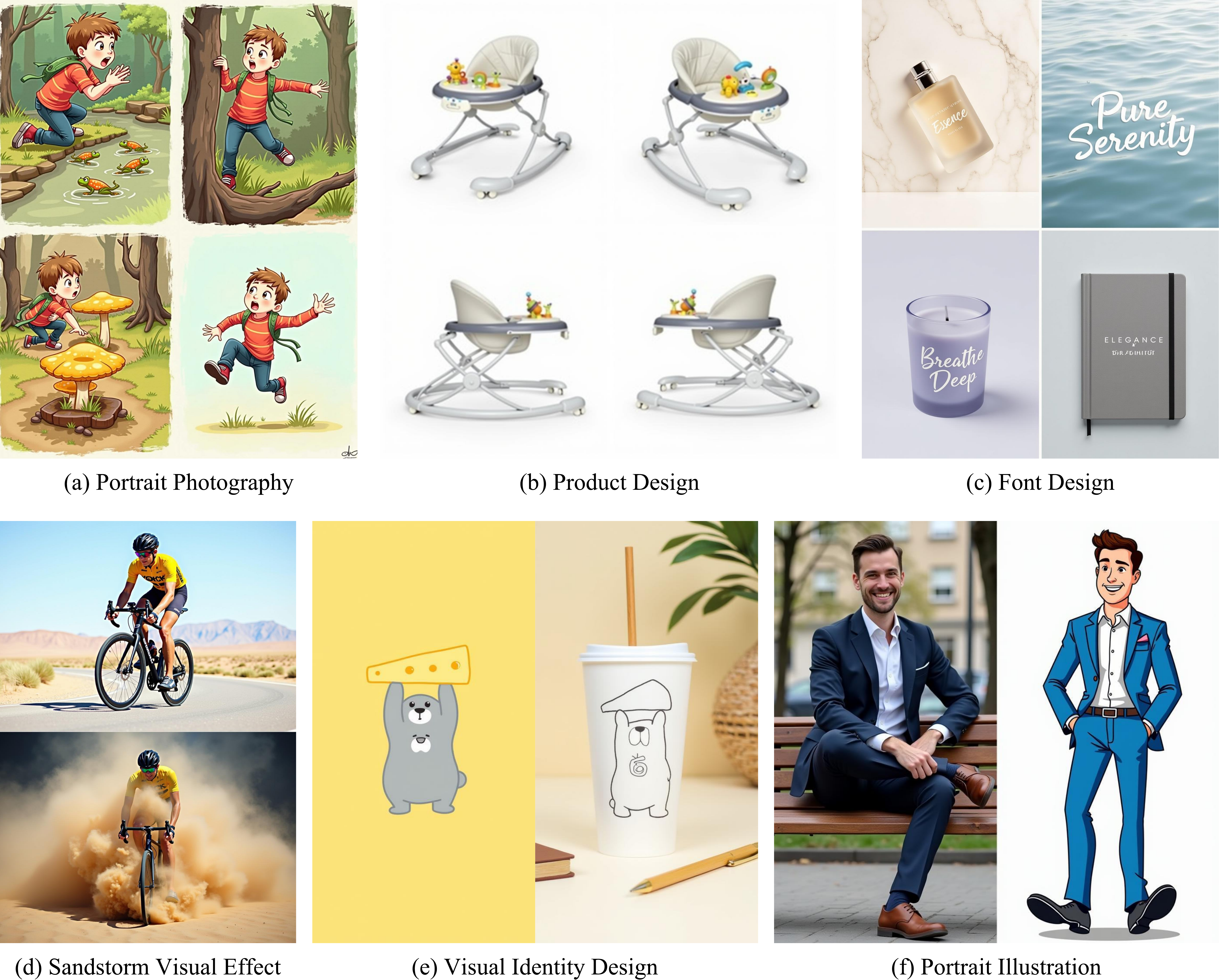}

    \caption{\textbf{FLUX Text-to-Image Generation Examples.} Examples of text-to-image generation across six tasks using \texttt{FLUX.1-dev}, highlighting the creation of multi-panel images with varied relational attributes. Key observations include: (1) The original text-to-image model can already generate multi-panel outputs with coherent consistency in identity, style, lighting, and font, though some minor imperfections remain. (2) \texttt{FLUX.1-dev} shows a strong capability in interpreting combined prompts that describe multiple panels, as further detailed in Appendix \ref{appendix:prompts}.}
    \label{fig:fig3}
\end{figure}

\section{Introduction}
\label{sec:introduction}

The advent of text-to-image models has significantly advanced the field of visual content generation, enabling the creation of high-fidelity images from textual descriptions \citep{ramesh2021zero,ramesh2022hierarchical,esser2021taming,rombach2022high,saharia2022photorealistic,betker2023improving,podell2023sdxl,esser2024scaling,baldridge2024imagen,blackforestlabs_flux_2024}. Numerous methods now offer enhanced control over various image attributes, allowing for finer adjustments during generation \citep{zhang2023adding,ye2023ip,huang2023composer,ruiz2023dreambooth,wang2024instantid,hertz2024style}. Despite these strides, adapting text-to-image models to a broad spectrum of generative tasks—particularly those requiring coherent image sets with complex intrinsic relationships—remains a open challenge. In this work, we introduce a task-agnostic framework designed to adapt text-to-image models to diverse generative tasks, aiming to provide a universal solution for versatile and controllable image generation.

Recent efforts, such as the Group Diffusion Transformers (GDT) framework \citep{huang2024group}, have explored reformulating visual generative tasks as a group generation problem. In this context, a set of images with arbitrary intrinsic relationships is generated simultaneously within a single denoising diffusion process, optionally conditioned on another set of images. The core idea of GDT involves concatenating attention tokens across images—both the conditional ones and those to be generated—while ensuring that the tokens of each image attend exclusively to their corresponding text tokens. This approach allows the model to adapt to multiple tasks in a task-agnostic, zero-shot manner without any fine-tuning or gradient updates.

However, despite its innovative architecture, GDT exhibits relatively low generation fidelity, often underperforming compared to the original pretrained text-to-image models. This limitation prompts a re-examination of the underlying assumptions and methodologies employed in adapting text-to-image models for complex generative tasks.

In this work, we make a pivotal assumption: \textbf{text-to-image models inherently possess in-context generation capabilities}. To validate this, we directly apply existing text-to-image models to a variety of tasks that require generating sets of images with diverse relationships. As illustrated in Figure \ref{fig:fig3}, using the \texttt{FLUX.1-dev} model \citep{blackforestlabs_flux_2024} as an example, we observe that the model can already perform different tasks, albeit with some imperfections. It maintains consistent attributes such as subject identities, styles, lighting conditions, and color palettes while modifying other aspects like poses, 3D orientations, and layouts. Moreover, the model demonstrates the ability to interpret and follow descriptions of multiple images within a single merged prompt, as detailed in Appendix \ref{appendix:prompts}.

These surprising findings lead us to several key insights:

\begin{enumerate}
    \item \textbf{Inherent In-Context Learning}: Text-to-image models already possess in-context generation abilities. By appropriately triggering and enhancing this capability, we can leverage it for complex generative tasks.
    
    \item \textbf{Model Reusability Without Architectural Modifications}: Since text-to-image models can interpret merged captions, we can reuse them for in-context generation without any changes to their architecture. This involves simply altering the input data rather than modifying the model itself.
    
    \item \textbf{Efficiency with Minimal Data and Computation}: High-quality results can be achieved without large datasets or prolonged training times. Small, high-quality datasets coupled with minimal computational resources may be sufficient.
\end{enumerate}

Building upon these insights, we design an extremely simple yet effective pipeline for adapting text-to-image models to diverse tasks. Our approach contrasts with GDT in the following ways:

\begin{enumerate}
    \item \textbf{Image Concatenation}: We concatenate a set of images into a single large image instead of concatenating attention tokens. This method is approximately equivalent to token concatenation in diffusion transformers (DiTs), disregarding differences introduced by the Variational Autoencoder (VAE) component.
    
    \item \textbf{Prompt Concatenation}: We merge per-image prompts into one long prompt, enabling the model to process and generate multiple images simultaneously. This differs from the GDT approach, where each image’s tokens cross-attend exclusively to its text tokens.
    
    \item \textbf{Minimal Fine-Tuning with Small Datasets}: Instead of performing large-scale training on hundreds of thousands of samples, we fine-tune a Low-Rank Adaptation (LoRA) of the model using a small set of just $20\sim 100$ image sets. This approach significantly reduces the computational resources required and largely preserves the original text-to-image model's knowledge and in-context capabilities.
\end{enumerate}

The resulting model is remarkably simple, requiring no modifications to the original text-to-image models. Adaptation is achieved solely by adjusting a small set of tuning data according to specific task needs. To support image-conditional generation, we employ a straightforward technique: we mask one or multiple images in the concatenated large image and prompt the model to inpaint them using the remaining images. We directly utilize SDEdit \citep{meng2021sdedit} for this purpose.

Despite its simplicity, we find that our method can adapt to a diverse array of tasks with high quality. Figures \ref{fig:fig1} and \ref{fig:fig2} illustrate example outputs for various tasks, while Figures \ref{fig:fig4}--\ref{fig:fig12} present more specific cases by task, and Figures \ref{fig:fig13} and \ref{fig:fig14} demonstrate image-conditional generation results. While our approach requires task-specific tuning data, the overall framework and pipeline remain task-agnostic, allowing adaptation to a wide variety of tasks without modifying the original model architecture. This combination of minimal data requirements and broad applicability offers a powerful tool for the generative community, designers, and artists. We acknowledge that developing a fully unified generation system remains an open challenge and leave it as future work. To facilitate further research, we release our data, models, and training configurations at the project page\footnote{Project page: \url{https://ali-vilab.github.io/In-Context-Lora-Page/}}.

\section{Related Work}
\label{sec:related_work}

\subsection{Task-Specific Image Generation}

Text-to-image models have achieved remarkable success in generating high-fidelity images from complex textual prompts \citep{ramesh2021zero,ramesh2022hierarchical,esser2021taming,rombach2022high,saharia2022photorealistic,betker2023improving,podell2023sdxl,chen2023pixart,esser2024scaling,baldridge2024imagen,blackforestlabs_flux_2024}. However, they often lack fine-grained controllability over specific attributes of the generated images. To address this limitation, numerous works have been proposed to enhance control over aspects such as layouts \citep{zheng2023layoutdiffusion,huang2023composer}, poses \citep{zhang2023adding}, identities \cite{huang2023composer,ye2023ip,li2023photomaker,wang2024instantid}, color palettes \citep{huang2023composer}, styles \cite{hertz2024style,huang2023composer}, regions \citep{meng2021sdedit,lugmayr2022repaint,xie2022smartbrushtextshapeguided,huang2023composer}, handles \citep{pan2023drag,shi2023dragdiffusion,liu2024drag}, distorted images \citep{saharia2022image,kawar2022denoising,xia2023diffirefficientdiffusionmodel,li2023diffusion} and lighting conditions \citep{zhang2023adding}. Some methods even support the simultaneous generation of multiple images, akin to our approach \citep{zhou2024storydiffusion,liu2024intelligent,yang2024seed,chern2024anole,huang2024group}.

Despite these advancements, these models typically employ task-specific architectures and pipelines, limiting their flexibility and generalizability. Each architecture is tailored to individual tasks, and the capabilities developed for one task are not easily composable or extendable to arbitrary new tasks. This contrasts with recent progress in natural language processing \citep{radford2019language,brown2020language,touvron2023llama,touvron2023llama2,dubey2024llama,geminiteam2024geminifamilyhighlycapable}, where models are designed to perform multiple tasks within a single architecture and can generalize beyond the tasks they were explicitly trained on.

\subsection{Task-Agnostic Image Generation}

To overcome the constraints of task-specific models, recent research has aimed at creating task-agnostic frameworks that support multiple controllable image generation tasks within a single architecture \citep{ge2023making, zhou2024transfusionpredicttokendiffuse, sheynin2024emu, sun2024generative, wang2024emu3}. For example, Emu Edit \citep{sheynin2024emu} integrates a broad array of image editing functions, while models like Emu2 \citep{sun2024generative}, Emu3 \citep{wang2024emu3}, TransFusion \citep{zhou2024transfusionpredicttokendiffuse}, Show-o \citep{xie2024showo}, and OmniGen \citep{xiao2024omnigen} perform diverse tasks, from procedural drawing to subject-driven generation, within a unified model. Emu3 further extends this capability by supporting text, image, and video generation under a single framework. These works represent substantial advancements in the unified or task-agnostic generation.

In contrast to these models, we propose that existing text-to-image architectures already possess inherent \textit{in-context capabilities}. This eliminates the need for developing new architectures, and enabling high-quality generation with minimal additional data and computational resources. Our approach not only enhances efficiency but also delivers superior generation quality across a wide array of tasks.

\section{Method}
\label{sec:method}

\subsection{Problem Formulation}

Following the approach of Group Diffusion Transformers \citep{huang2024group}, we frame most image generation tasks as producing a set of $n \ge 1$ images, conditioned on another set of $m \ge 0$ images and $(n + m)$ text prompts. This formulation encompasses a broad range of academic tasks, such as image translation, style transfer, pose transfer, and subject-driven generation, as well as practical applications like picture book creation, font design and transfer, storyboard generation, and more \citep{huang2024group}. The correlations among both the conditional images and the generated images are implicitly maintained through the per-image prompts.

Our approach slightly modifies this framework by using a single consolidated prompt for the entire image set. This prompt typically begins with an overall description of the image set, followed by individual prompts for each image. This unified prompt design is more compatible with existing text-to-image models and allows the overall description to naturally convey the task’s intent, much like \textbf{how clients communicate design requirements to artists}.

\subsection{Group Diffusion Transformers}

We begin with the base framework, Group Diffusion Transformers (GDT) \citep{huang2024group}. In GDT, a set of images are generated simultaneously within a single diffusion process by concatenating attention tokens across images in each Transformer self-attention block. This approach enables each image to "see" and interact with all other images in the set. Text conditioning is introduced by having each image attend to its corresponding text embeddings, allowing it to access both the content of other images and relevant text guidance.

GDT is trained on hundreds of thousands of image sets, enabling it to generalize across tasks in a zero-shot manner.

\subsection{In-Context LoRA}

Although GDT demonstrates zero-shot task adaptability, its generation quality falls short, often underperforming compared to baseline text-to-image models. We propose enhancements to improve this framework.

Our starting point is the assumption that base text-to-image models inherently possess some in-context generation capabilities for diverse tasks, even if quality varies. This is supported by results in Figure \ref{fig:fig3}, where the model effectively generates multiple images (sometimes with conditions) across different tasks. Based on this insight, extensive training on large datasets is unnecessary; we can instead activate the model’s in-context abilities with carefully curated, high-quality image sets.

Another observation is that text-to-image models can generate coherent multi-panel images from a single prompt containing descriptions of multiple panels (see Figure \ref{fig:fig3} Appendix \ref{appendix:prompts}). Thus, we can simplify the architecture by using consolidated image prompts instead of requiring each image to attend exclusively to its respective text tokens. This allows us to reuse the original text-to-image architecture without any structural modifications.

Our final framework design generates a set of images simultaneously by directly concatenating them into a single large image during training, while consolidating their captions into one merged prompt with an overarching description and clear guidance for each panel. After generating the image set, we split the large image into individual panels. Furthermore, since text-to-image models already demonstrate in-context capabilities, we don’t fine-tune the entire model. Instead, we apply Low-Rank Adaptation (LoRA) on a small set of high-quality data to trigger and enhance these capabilities.

To support conditioning on an additional set of images, we employ SDEdit, a training-free method, to inpaint a set of images based on an unmasked set, all concatenated within a single large image.

\section{Experiments}

\subsection{Implementation Details}

We build our approach on the \texttt{FLUX.1-dev} text-to-image model \citep{blackforestlabs_flux_2024} and train an In-Context LoRA specifically for our tasks. We select a range of practical tasks, including storyboard generation, font design, portrait photography, visual identity design, home decoration, visual effects, portrait illustration, and PowerPoint template design, among others. For each task, we collect 20 to 100 high-quality image sets from the internet. Each set is concatenated into a single composite image, and captions for these images are generated using Multi-modal Large Language Models (MLLMs), starting with an overall summary followed by detailed descriptions for each image. Training is conducted on a single A100 GPU for 5,000 steps with a batch size of 4 and a LoRA rank of 16. For inference, we employ 20 sampling steps with a guidance scale of 3.5, matching the distillation guidance scale of \texttt{FLUX.1-dev}. For image-conditional generation, SDEdit is applied to mask images intended for generation, enabling inpainting based on the surrounding images.

\subsection{Results}

We present qualitative results demonstrating the versatility and quality of our model across various tasks. Given the wide diversity of tasks, we defer a unified quantitative benchmark and evaluation to future work.

\subsubsection{Reference-Free Image-Set Generation}

\begin{figure}
    \centering
    \vspace{-12pt}

    \makebox[\textwidth]{%
        \hspace{-50pt}
        % \fbox{\rule{0pt}{0.95\textheight}\rule{1.1\textwidth}{0pt}} % 用于占位的矩形框
        \includegraphics[width=1.1\textwidth]{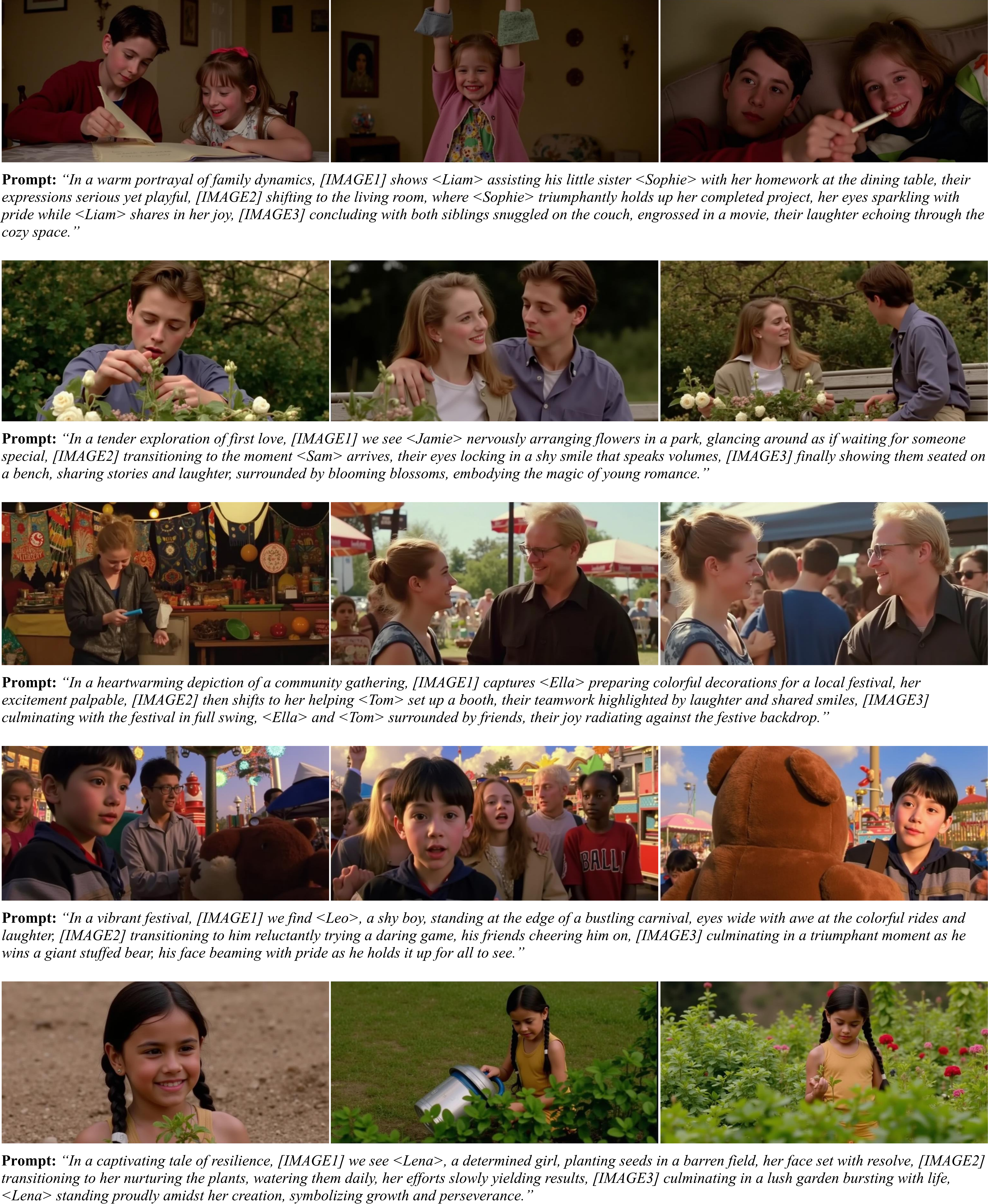}
        \hspace{-50pt}
    }

    \caption{\textbf{Film Storyboard Generation.} Each set of three images is generated simultaneously using In-Context LoRA. A placeholder \textbf{character name} wrapped in "\textit{<}" and "\textit{>}" uniquely references the character's identity across the images, ensuring consistent portrayal throughout the storyboard.}
    \label{fig:fig4}
\end{figure}

\begin{figure}
    \centering
    \vspace{-12pt}

    \makebox[\textwidth]{%
        \hspace{-45pt}
        % \fbox{\rule{0pt}{0.95\textheight}\rule{1.1\textwidth}{0pt}} % 用于占位的矩形框
        \includegraphics[width=1.09\textwidth]{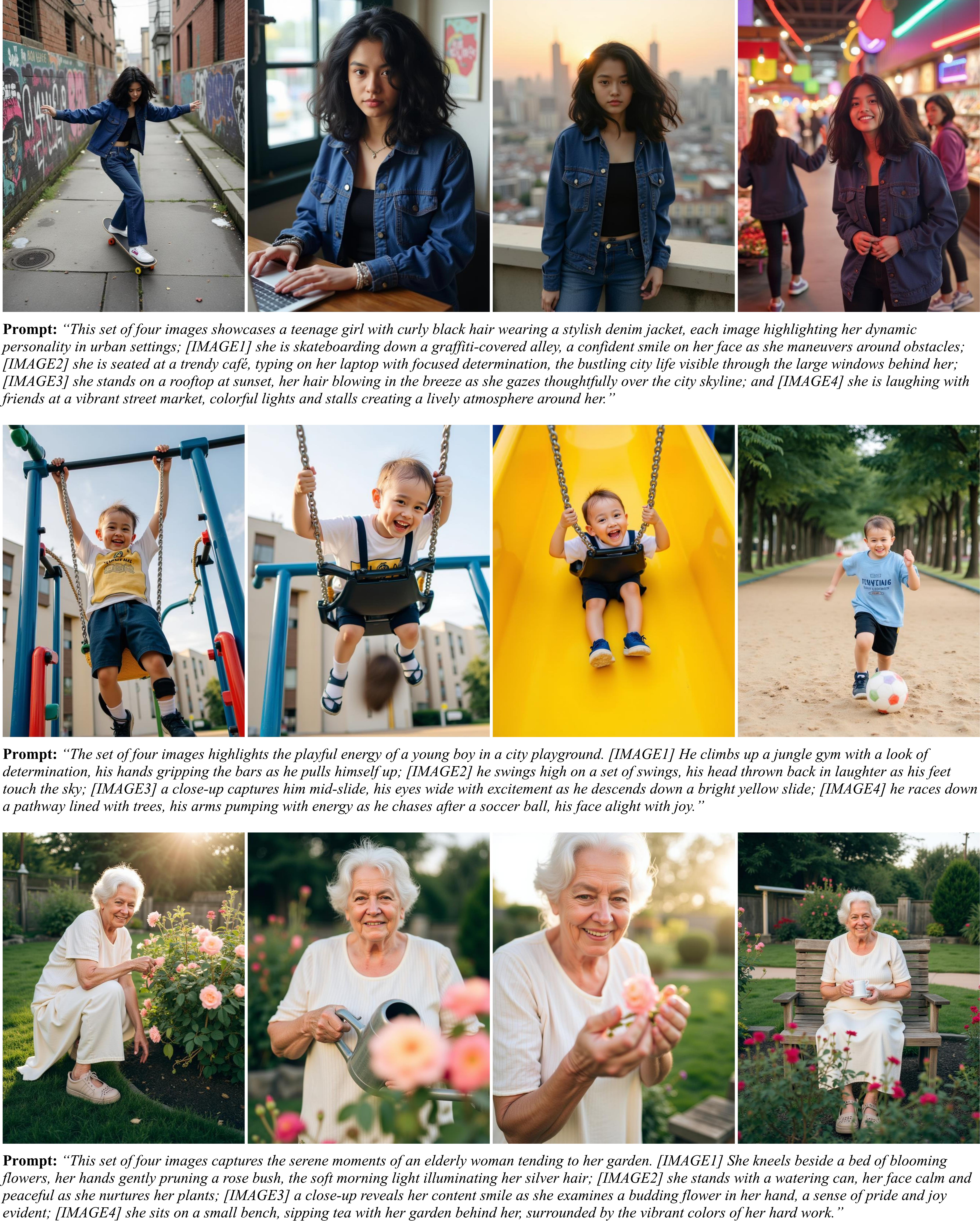}
        \hspace{-45pt}
    }

    \caption{\textbf{Portrait Photography.} Each set of four images is generated simultaneously using In-Context LoRA. Consistent subject identities are maintained across all images within each set, as illustrated in the figure.}
    \label{fig:fig5}
\end{figure}

\begin{figure}
    \centering
    \vspace{-12pt}

    \makebox[\textwidth]{%
        \hspace{-50pt}
        % \fbox{\rule{0pt}{0.95\textheight}\rule{1.1\textwidth}{0pt}} % 用于占位的矩形框
        \includegraphics[width=1.1\textwidth]{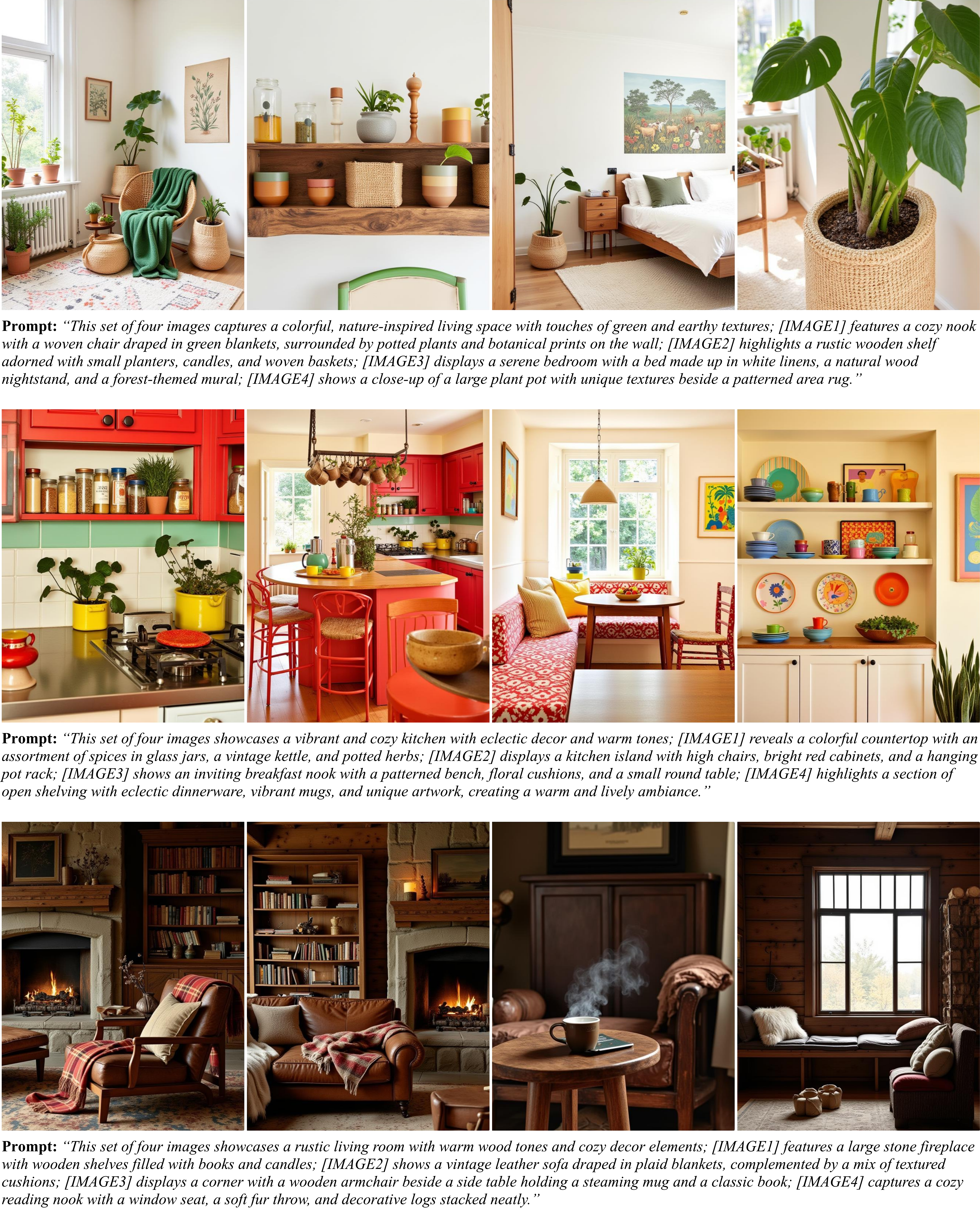}
        \hspace{-50pt}
    }

    \caption{\textbf{Home Decoration.} Each set of four images is generated simultaneously using In-Context LoRA, showcasing a consistent decoration style across all images within each set.}
    \label{fig:fig6}
\end{figure}

\begin{figure}
    \centering
    \vspace{-15pt}

    \makebox[\textwidth]{%
        \hspace{-50pt}
        % \fbox{\rule{0pt}{0.95\textheight}\rule{1.1\textwidth}{0pt}} % 用于占位的矩形框
        \includegraphics[width=1.1\textwidth]{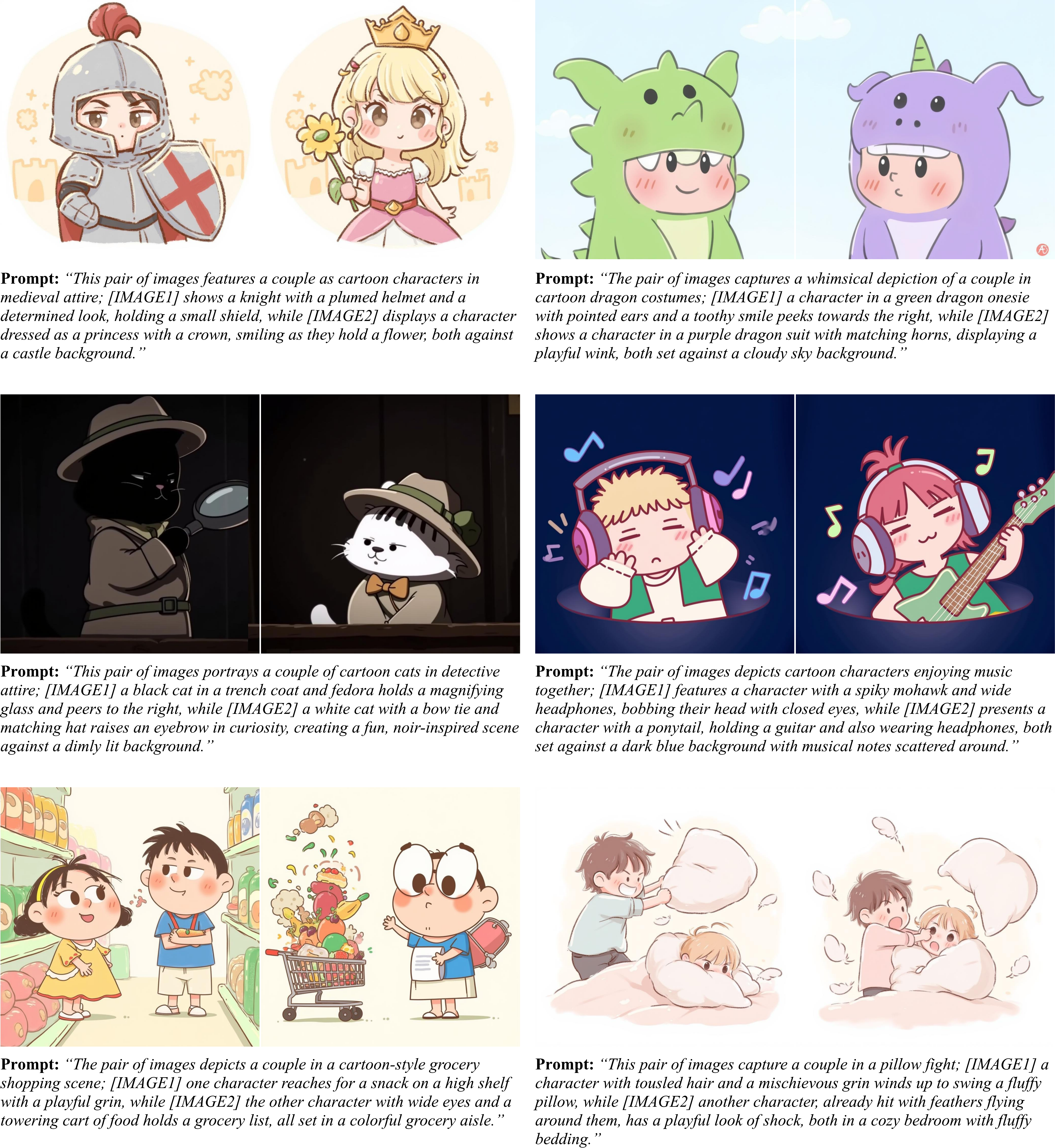}
        \hspace{-50pt}
    }

    \caption{\textbf{Couple Profile Generation.} Each pair of images is generated simultaneously using In-Context LoRA, maintaining consistent style and identity features across both images in each set.}
    \label{fig:fig7}
\end{figure}

\begin{figure}
    \centering
    \vspace{-15pt}

    \makebox[\textwidth]{%
        \hspace{-50pt}
        % \fbox{\rule{0pt}{0.95\textheight}\rule{1.1\textwidth}{0pt}} % 用于占位的矩形框
        \includegraphics[width=1.1\textwidth]{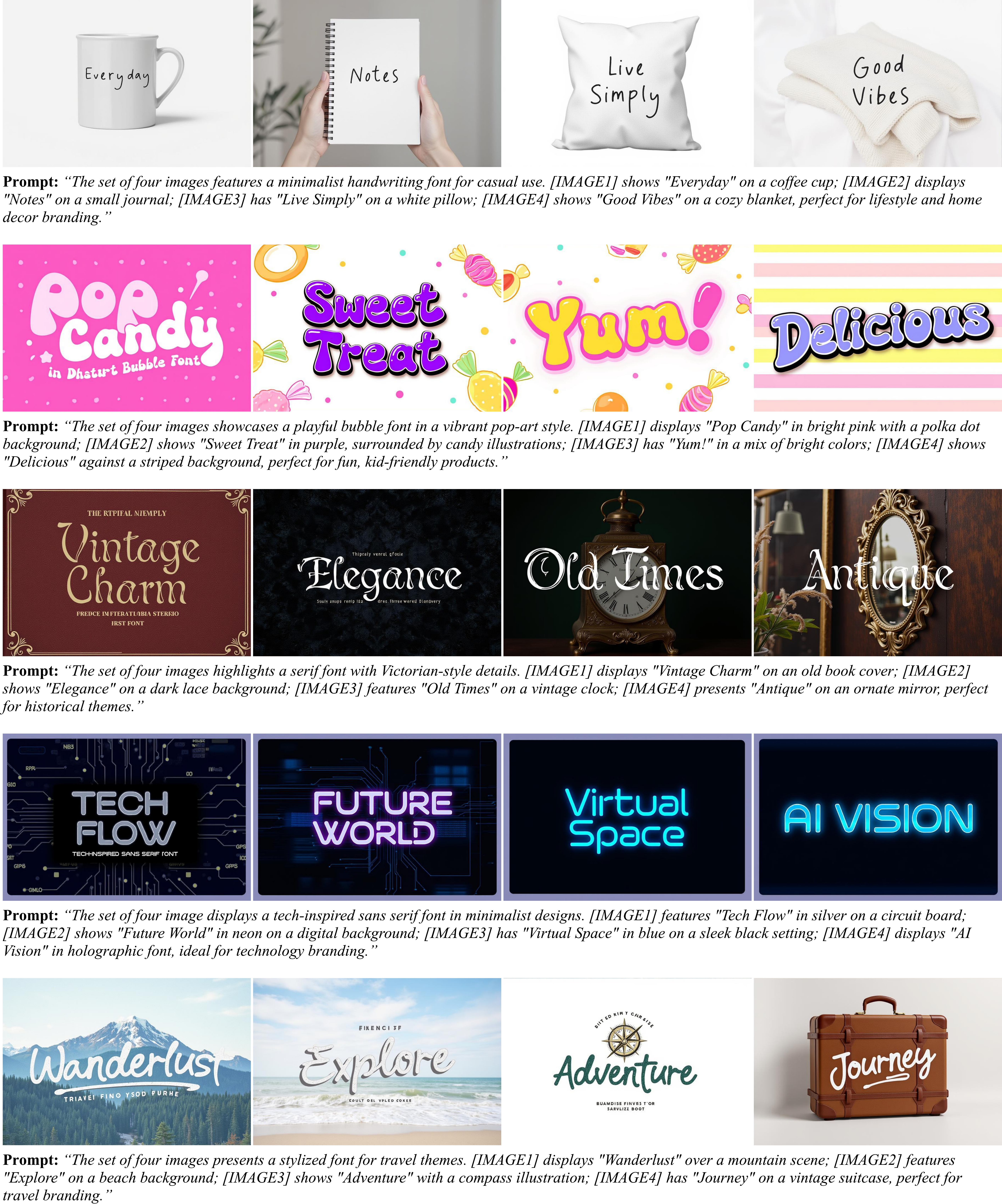}
        \hspace{-50pt}
    }

    \caption{\textbf{Font Design.} Each set of four images is generated simultaneously using In-Context LoRA, ensuring a consistent font style throughout all images in each set.}
    \label{fig:fig8}
\end{figure}

\begin{figure}
    \centering
    \vspace{-12pt}

    \makebox[\textwidth]{%
        \hspace{-55pt}
        % \fbox{\rule{0pt}{0.95\textheight}\rule{1.1\textwidth}{0pt}} % 用于占位的矩形框
        \includegraphics[width=1.15\textwidth]{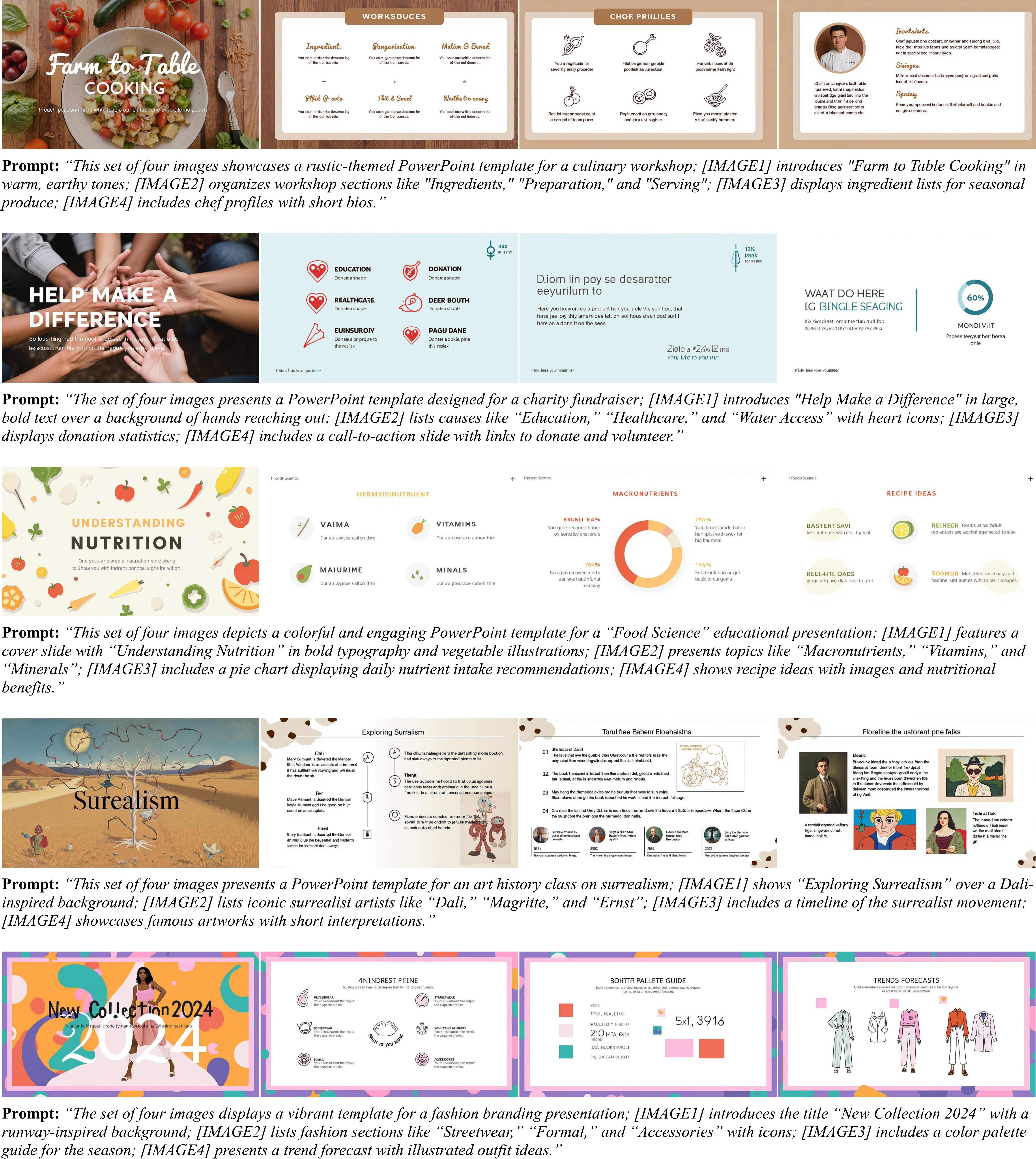}
        \hspace{-55pt}
    }

    \caption{\textbf{PowerPoint Template Design.} Each set of four images is generated simultaneously using In-Context LoRA, achieving a cohesive and unified presentation style across all slides within each set.}
    \label{fig:fig9}
\end{figure}

\begin{figure}
    \centering
    \vspace{-12pt}

    \makebox[\textwidth]{%
        \hspace{-55pt}
        % \fbox{\rule{0pt}{0.95\textheight}\rule{1.1\textwidth}{0pt}} % 用于占位的矩形框
        \includegraphics[width=1.15\textwidth]{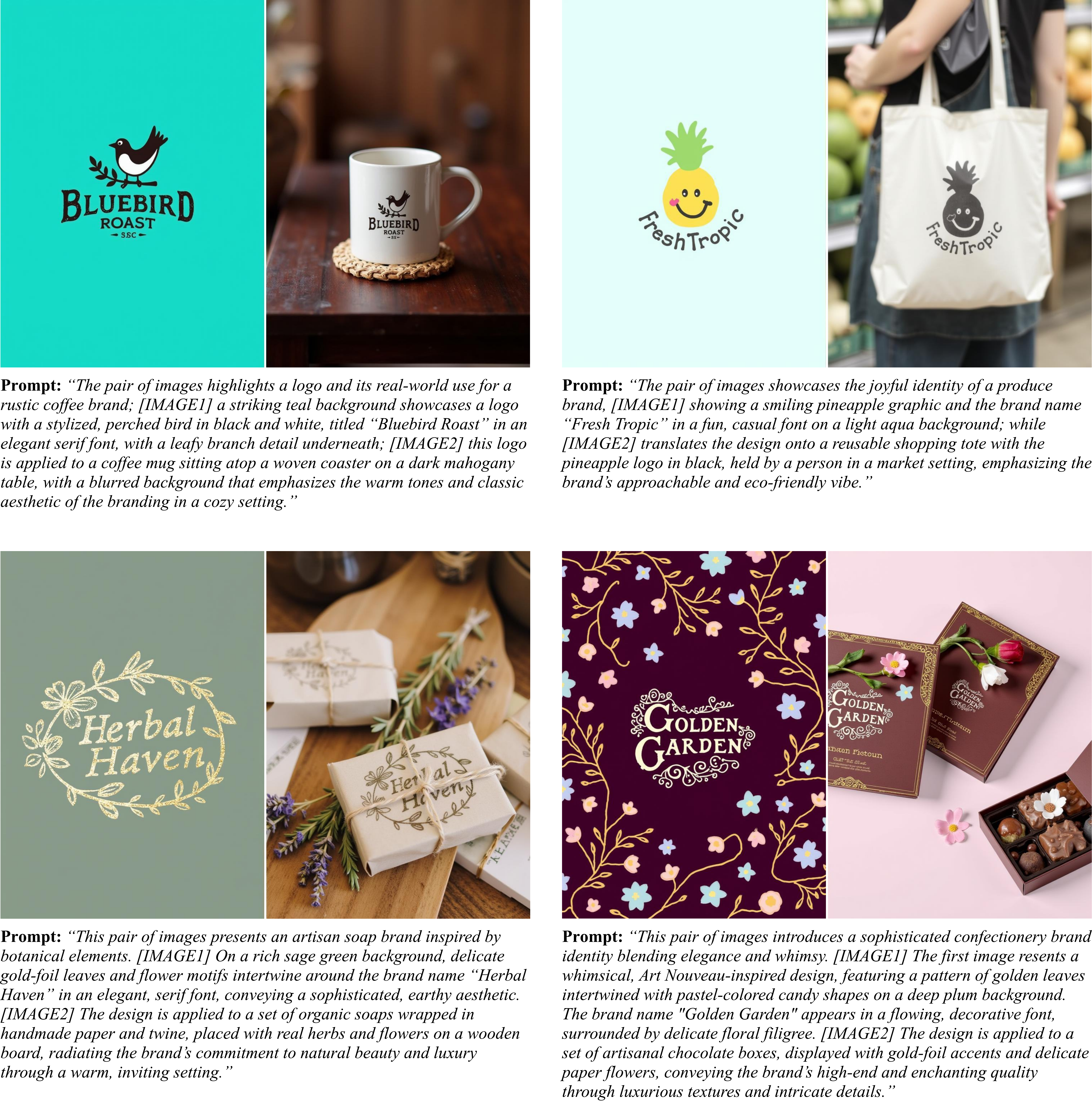}
        \hspace{-55pt}
    }

    \caption{\textbf{Visual Identity Design.} Each pair of images is generated simultaneously using In-Context LoRA, ensuring a cohesive and consistent visual identity across both images in each pair.}
    \label{fig:fig10}
\end{figure}

\begin{figure}
    \centering
    \vspace{-15pt}

    \makebox[\textwidth]{%
        \hspace{-50pt}
        % \fbox{\rule{0pt}{0.95\textheight}\rule{1.1\textwidth}{0pt}} % 用于占位的矩形框
        \includegraphics[width=1.1\textwidth]{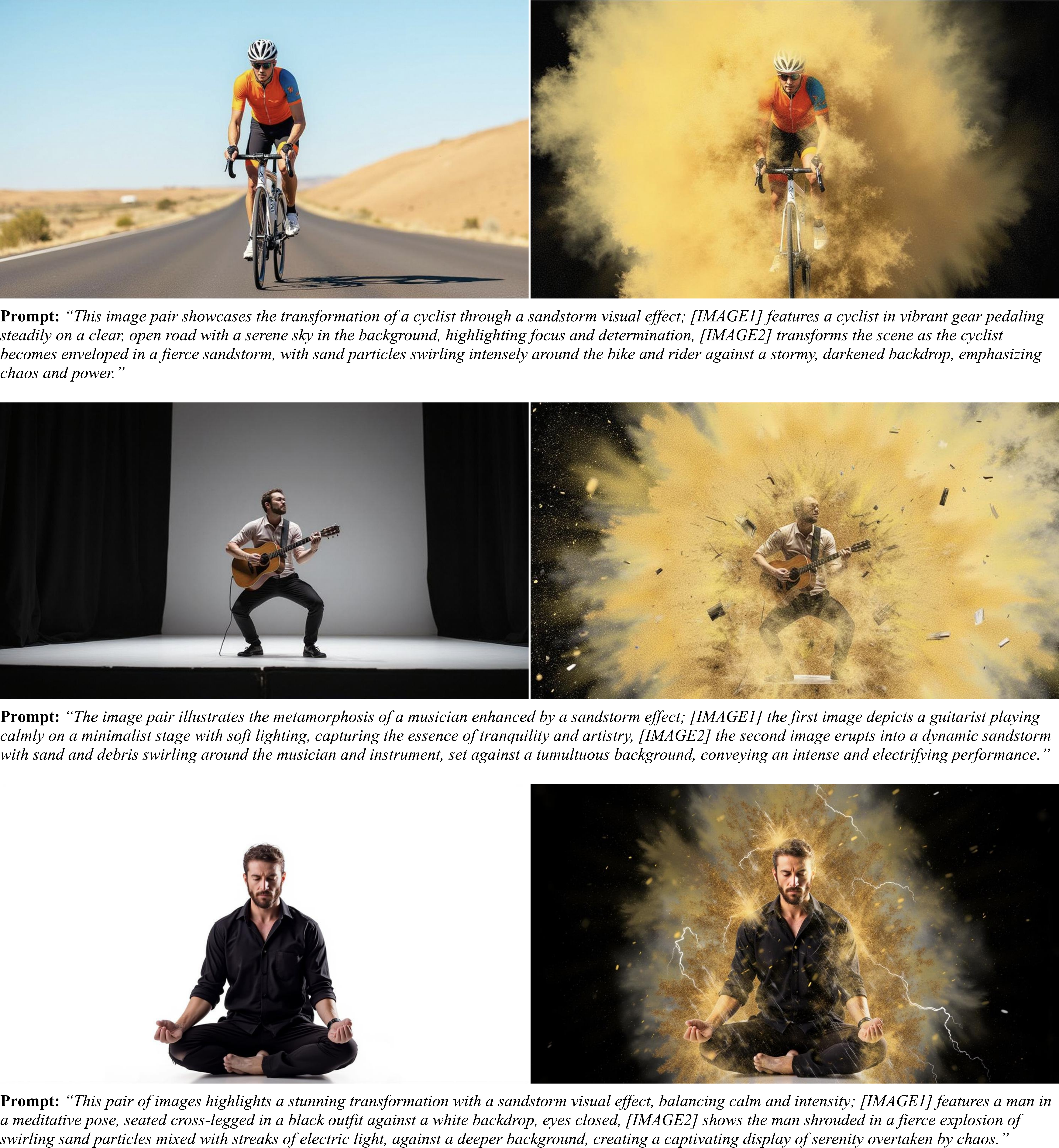}
        \hspace{-50pt}
    }

    \caption{\textbf{Sandstorm Visual Effect.} Each pair of images is generated using In-Context LoRA, demonstrating strong consistency between the "before" and "after" sandstorm effect images. For examples of image-conditional generation, please refer to Figure \ref{fig:fig13}.}
    \label{fig:fig11}
\end{figure}

\begin{figure}
    \centering
    \vspace{-15pt}

    \makebox[\textwidth]{%
        \hspace{-30pt}
        % \fbox{\rule{0pt}{0.95\textheight}\rule{1.1\textwidth}{0pt}} % 用于占位的矩形框
        \includegraphics[width=1.05\textwidth]{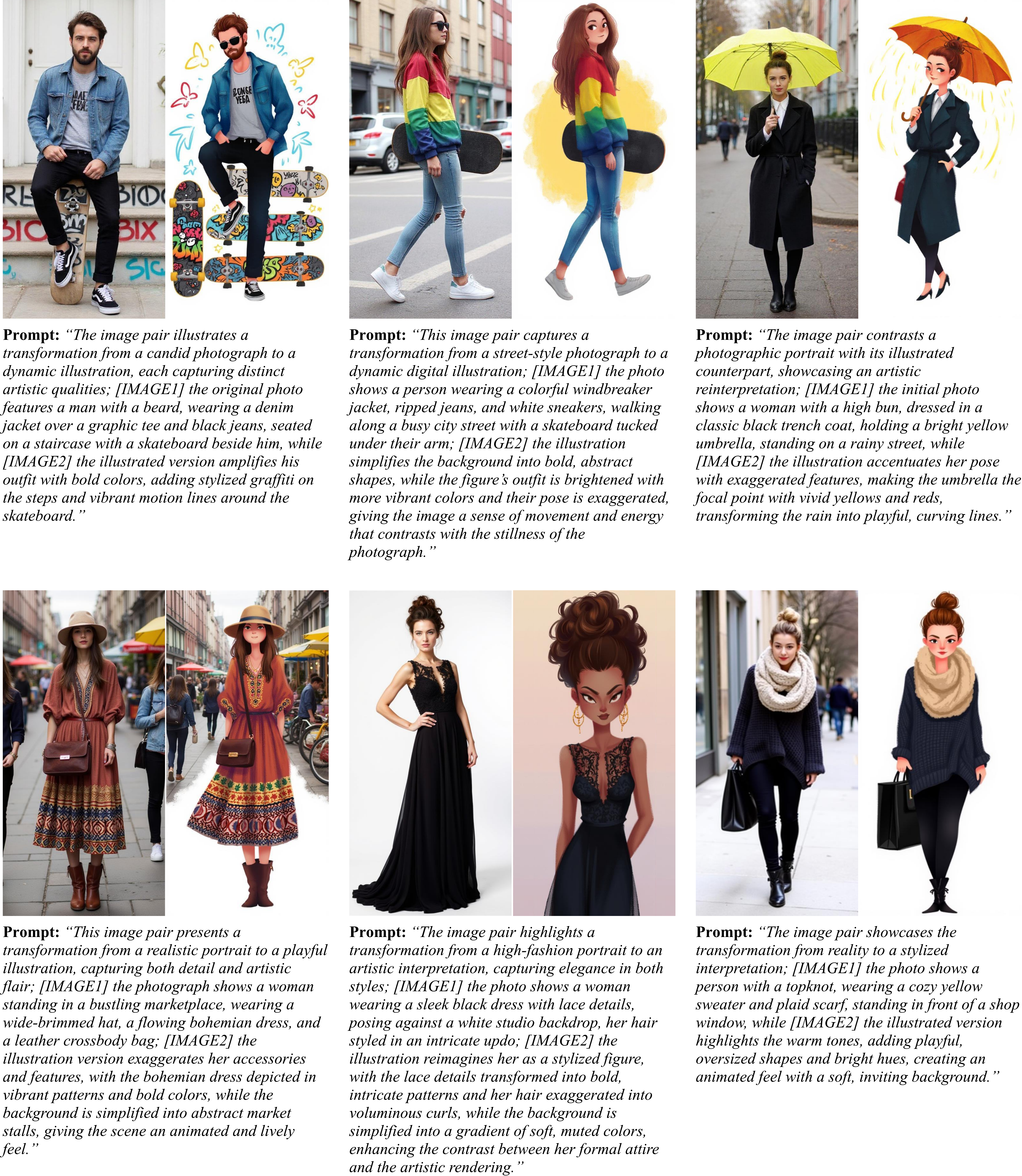}
        \hspace{-30pt}
    }

    \caption{\textbf{Portrait Illustration.} Each pair of images is generated using In-Context LoRA, maintaining consistent identity, clothing, expression, similar pose, and atmosphere between the "before" and "after" illustration versions. Rather than directly copying the original photo, the illustration artistically enhances key features, adding expressive emphasis. For additional examples of image-conditional generation, please refer to Figure \ref{fig:fig13}.}
    \label{fig:fig12}
\end{figure}

In this setting, image sets are generated solely from text prompts, with no additional image input. Examples from a range of tasks are presented in Figures \ref{fig:fig4}--\ref{fig:fig12}. Our method achieves high-quality results across a spectrum of image-set generation tasks.

\subsubsection{Reference-Based Image-Set Generation}

\begin{figure}
    \centering
    \vspace{-15pt}

    \makebox[\textwidth]{%
        \hspace{-30pt}
        % \fbox{\rule{0pt}{0.95\textheight}\rule{1.1\textwidth}{0pt}} % 用于占位的矩形框
        \includegraphics[width=1.05\textwidth]{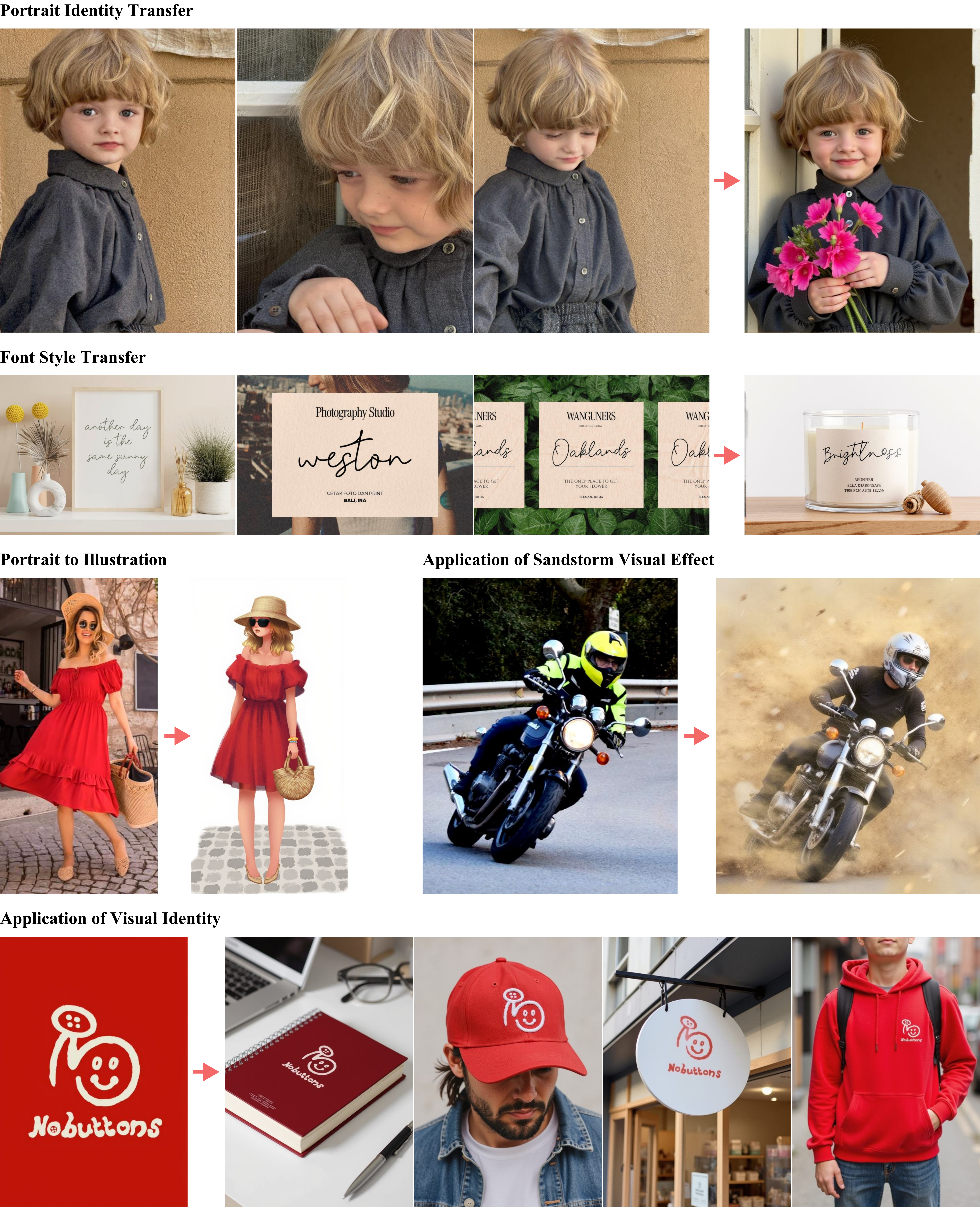}
        \hspace{-30pt}
    }

    \caption{\textbf{Image-Conditional Generation.} Examples of image-conditional generation using In-Context LoRA across multiple tasks with training-free SDEdit. In some instances, such as the \textit{Application of Sandstorm Visual Effect} case, inconsistencies may arise between input and output images, including changes in the motor driver's identity and attire. Addressing these inconsistencies is left for future work.}
    \label{fig:fig13}
\end{figure}

\begin{figure}
    \centering
    \includegraphics[width=0.9\textwidth]{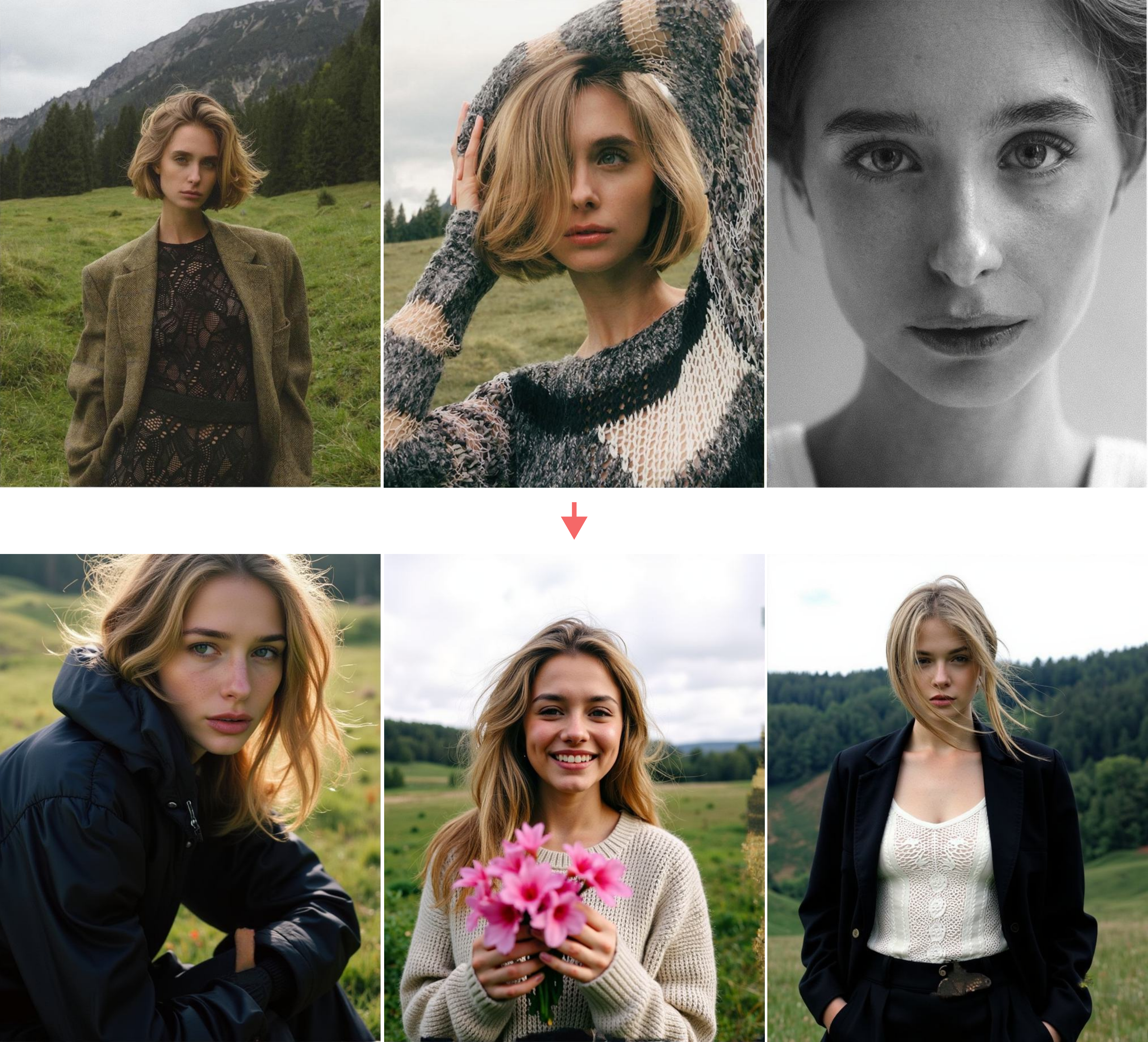}

    \caption{\textbf{Failure Cases of Image-Conditional Generation.} Examples of portrait identity transfer failure using In-Context LoRA with SDEdit. We observe that SDEdit for In-Context LoRA tends to be unstable, often failing to preserve identity. This may stem from a discrepancy between SDEdit’s \textbf{unidirectional dependency} on input-to-output mapping and the \textbf{bidirectional nature} of In-Context LoRA training. Addressing this issue is left for future work.}

    \label{fig:fig14}
\end{figure}

In this setting, image sets are generated using both a text prompt and an input image set (with at least one reference image). SDEdit is applied to mask certain images, enabling inpainting based on the remaining ones. Results for image-conditioned generation are presented in Figure \ref{fig:fig13}, with common failure cases shown in Figure \ref{fig:fig14}. Although effective across multiple tasks, visual consistency across images is sometimes lower compared to text-conditioned generation. This discrepancy may result from SDEdit’s unidirectional dependency between masked and unmasked images, whereas text-only generation allows bidirectional dependencies among images, enabling mutual adjustment of conditions and outputs. This suggests potential for improvement, such as incorporating a trainable inpainting method, which we leave for future exploration.

\newpage

\bibliographystyle{unsrtnat}
\bibliography{references}

\appendix

\section{Prompts of Figure \ref{fig:fig3}}
\label{appendix:prompts}

Below are the detailed prompts used for each subfigure in Figure \ref{fig:fig3}:

\paragraph{(a) Portrait Photography.} This four-panel image captures a young boy's adventure in the woods, expressing curiosity and wonder. [TOP-LEFT] He crouches beside a stream, peering intently at a group of frogs jumping along the rocks, his face full of excitement; [TOP-RIGHT] he climbs a low tree branch, arms stretched wide as he balances, a big grin on his face; [BOTTOM-LEFT] a close-up shows him kneeling in the dirt, inspecting a bright yellow mushroom with fascination; [BOTTOM-RIGHT] the boy runs through a clearing, his arms spread out like airplane wings, lost in the thrill of discovery.

\paragraph{(b) Product Design.} The image showcases a modern and multifunctional baby walker designed for play and growth, featuring its versatility and attention to detail; [TOP-LEFT] the first panel highlights the walker’s sleek form with several interactive toys on the tray, focusing on its overall structure and functionality, [TOP-RIGHT] the second panel provides a side view emphasizing the built-in lighting around the play tray, illustrating both style and safety features, [BOTTOM-LEFT] the third panel displays a rear view of the walker showcasing the comfortable seat and adjustable design elements, underlining comfort and adaptability, [BOTTOM-RIGHT] while the final panel offers a close-up of the activity center with various colorful toys, capturing its playful appeal and engagement potential.

\paragraph{(c) Font Design.} The four-panel image emphasizes the versatility of a minimalist sans-serif font across various elegant settings: [TOP-LEFT] displays the word “Essence” in muted beige, featured on a luxury perfume bottle with a marble backdrop; [TOP-RIGHT] shows the phrase “Pure Serenity” in soft white, set against an image of serene, rippling water; [BOTTOM-LEFT] showcases “Breathe Deep” in pale blue, printed on a calming lavender candle, evoking a spa-like atmosphere; [BOTTOM-RIGHT] features “Elegance Defined” in charcoal gray, embossed on a sleek hardcover notebook, emphasizing sophistication and style.

\paragraph{(d) Sandstorm Visual Effect.} The two-panel image showcases a biker speeding through a desert landscape before and after a sandstorm effect, capturing a powerful transformation; [TOP] the first panel presents a biker riding along a dirt path, with the vast desert and blue sky stretching out behind them, conveying a sense of freedom and adventure, while [BOTTOM] the second panel introduces a violent sandstorm, with grains of sand swirling around the biker and partially obscuring the landscape, transforming the scene into a chaotic and thrilling visual spectacle.

\paragraph{(e) Visual Identity Design.} This two-panel image captures the essence of a visual identity design and its adaptable application, showcasing both the original concept and its practical derivative use; [LEFT] the left panel presents a bright and engaging graphic featuring a stylized gray koala character triumphantly holding a large wedge of cheese on a vibrant yellow background, using bold black outlines to emphasize the simplicity and playfulness of the design, while [RIGHT] the right panel illustrates the design's extension to everyday objects, where the same koala and cheese motif has been skillfully adapted onto a circular coaster with a softer yellow tone, accompanied by a matching mug bearing smaller graphics of the koala and cheese, both items resting elegantly on a minimalist white table, highlighting the versatility and cohesive appeal of the visual identity across different mediums.

\paragraph{(f) Portrait Illustration.} This two-panel image showcases a transformation from a photographic portrait to a playful illustration; [LEFT] the first panel displays a man in a navy suit, white shirt, and brown shoes, sitting on a wooden bench in an urban park, his hand resting casually on his lap; [RIGHT] the illustration panel transforms him into a cartoon-like character, with smooth lines and exaggerated features, including oversized shoes and a vibrant blue suit, set against a minimalist park backdrop, giving the scene a lively and humorous feel.

\end{document}